\documentclass{article}
\usepackage{spconf,amssymb,amsmath,graphicx}

\usepackage{float}
\usepackage{fixltx2e}
\usepackage{color}
\usepackage{colortbl}

\newcommand{\quotes}[1]{``#1''}

\title{SEMI-AUTOMATIC SEGMENTATION OF AUTOSOMAL DOMINANT POLYCYSTIC KIDNEYS USING RANDOM FORESTS}
\name{\small Kanishka Sharma$^{1,2}$,\thanks{Part of this work was funded by TranCYST Marie Curie Initial Training Networks project, within the 7th European Community Framework Programme (FP7-PEOPLE-MCA-ITN-317246).}Lo\"ic Peter$^{2}$,Christian Rupprecht$^{2,3}$,Anna Caroli$^{1}$,Lichao Wang$^{2}$,Andrea Remuzzi$^{1,4}$,Maximilian Baust$^{2}$,Nassir Navab$^{2,3}$}

 \address{$^{1}$ IRCCS - Istituto di Ricerche Farmacologiche \quotes{Mario Negri}, Bergamo, Italy \\
     $^{2}$ Computer Aided Medical Procedures, Technische Universit\"at M\"unchen, Germany\\
      $^{3}$ Computer Aided Medical Procedures, Johns Hopkins University, USA\\
      $^{4}$ University of Bergamo, Bergamo, Italy}

\begin{document}
%
\maketitle
\begin{abstract}
\textbf{This paper presents a method for 3D segmentation of kidneys from patients with autosomal dominant polycystic kidney disease (ADPKD) and severe renal insufficiency, using computed tomography (CT) data. 
ADPKD severely alters the shape of the kidneys due to non-uniform formation of cysts. 
As a consequence, fully automatic segmentation of such kidneys is very challenging. 
We present a segmentation method with minimal user interaction based on a random forest classifier. 
One of the major novelties of the proposed approach is the usage of geodesic distance volumes as additional source of information. 
These volumes contain the intensity weighted distance to a manual outline of the respective kidney in only one slice (for each kidney) of the CT volume. 
We evaluate our method qualitatively and quantitatively on 55 CT acquisitions using ground truth annotations from clinical experts.}
\end{abstract}
\section{Introduction}
\label{sec:intro}

Autosomal dominant polycystic kidney disease (ADPKD) is the most common inherited cystic kidney disease. 
It is characterized by progressive enlargement of the kidneys caused by sustained development and expansion of renal cysts \cite{grantham1996etiology}, eventually inducing end stage renal disease (ESRD) in majority of the patients. 
In addition, ADPKD is associated with a number of extra renal complications such as early hypertension, cerebral aneurysms, heart valve defects, and presence of cysts in other organs like the liver and pancreas.
For the purpose of diagnosis, Ultrasonography (US) is usually performed for pre-symptomatic identification of ADPKD, followed by Computed Tomography (CT) or Magnetic Resonance Image (MRI) acquisitions, offering higher resolution \cite{torres2007autosomal}.
So far, there are no existing proven treatments for ADPKD, therefore an effective disease-modifying drug would have important implications for patients. 
Development of efficient computational means for monitoring kidney expansion by reliably quantifying total kidney volumes (TKV) is of crucial importance for assessment of ADPKD progression and evaluating the efficacy of novel therapies. 
Therefore, fast, accurate and user-friendly segmentation methods are required for TKV computation in ADPKD. 
However, segmentation of pathological kidneys is not straightforward due to non-uniform renal cyst growth in ADPKD, leading to high variability in terms of organ shape and size. 
In addition to this, adjacent liver cysts, typically exhibiting similar intensity statistics as kidney cysts \cite{racimora2010segmentation}, further hamper accurate kidney delineation. 
In case of late stage ADPKD patients, such as those with severe renal insufficiency, hemorrhagic kidney cysts lead to extra complication of high intensity variability with respect to regular fluid-filled cysts. 
A fully automated segmentation approach allowing reliable and quick TKV estimation remains a challenging task. 
In this paper, we present a semi-automatic segmentation method based on a random forest classifier for extracting polycystic kidneys from CT volumes. 
Our method has been used on ADPKD patients with severe renal insufficiency, utilizing a large dataset of 55 contrast enhanced CT acquisitions. 

\section{RELATED WORK}
\label{sec:format}

In ADPKD clinical studies, commonly employed image based methods for TKV quantification are stereology \cite{bae2000volumetric} and manual segmentation.
Both methods tend to be time consuming, especially in case of high resolution acquisitions. 
Recently, a new method for estimating TKV for ADPKD kidneys has been proposed \cite{bae2013novel}, which does not require whole kidney segmentation, but only requires the segmentation of kidney section in its corresponding mid-slice. 
Another method by Irazabal \emph{et al.} \cite{irazabal2014imaging} describes an ellipsoid formula for TKV computation in ADPKD. 
Even though these methods allows fast quantification, they are likely to provide only a rough TKV estimation which could be useful for ADPKD clinics, but not sufficient for clinical trials that need more accurate measurements.\\
Besides these approaches, several semi-automatic kidney segmentation techniques have been proposed previously. 
Fei \emph{et al.} \cite{fei2006image} used a deformable model based on a spline parameterization for the segmentation of polycystic kidneys from transgenic mice, but the data sets used exhibited less severe morphological changes of the kidneys. 
Daum \emph{et al.} \cite{univis90709650} proposed a semi-automatic approach based on 3D random walks for the segmentation of polycystic kidneys on T2 weighted fat saturated MR acquisitions. 
This approach requires manual initialization by the user for determining connected non-labeled regions based on the probability of association to the user labeled pixels. 
Racimora \emph{et al.} \cite{racimora2010segmentation} investigated a semi-automatic approach using active contours and morphological operations. 
Mignani \emph{et al.} \cite{mignani2011assessment} used region growing segmentation with additional morphological operations and curvature based motion for assessment of TKV in T2 weighted fat saturated MRI images. 
An extension of their work was presented by Turco \emph{et al.} \cite{turco2015reliability}.
Another recent approach by Kline \emph{et al.} \cite{kline2015automatic}, describes automatic segmentation of follow-up T1 MR images from baseline acquisition. 
Their method necessarily requires a baseline segmentation initialization to segment follow-up MRI scans of ADPKD patients. 
It should be noted, however, that all these approaches focus on the segmentation of kidneys from MRI data and we are not aware of any semi-automatic method for segmenting ADPKD kidneys from CT data.
In this paper, we propose a semi-automatic segmentation method based on a random forest classifier. 
Random forests, or more generally decision forests, have gained considerable interest in medical image segmentation. 
As an example, regression forests have been used for the automatic detection and localization of multiple organs in MRI by Pauly \emph{et al.} \cite{pauly2011fast} or CT by Criminisi \emph{et al.} \cite{criminisi2013regression}. 
Focusing on kidney segmentation, Cuingnet \emph{et al.} \cite{cuingnet2012automatic} used an approach based on regression forests for segmenting normal kidneys. 
Kontschieder \emph{et al.} \cite{kontschieder2013geof} performed semantic image segmentation using geodesic distances as an additional criterion for the node splitting in order to ensure spatial compactness of the pixel clusters of each child node. 
Peter \emph{et al.} \cite{peter2015scale} presented scale adaptive segmentation approach on MR, 3D US and histological data using haar-like features sampled sequentially on a Random Forest classifier.
There are many more approaches based on random forests but an extensive review of all of them goes beyond the scope of this paper. 

\begin{figure}[htb]
	\centering
         \begin{tabular}{ccc}
         \includegraphics[height=32mm]{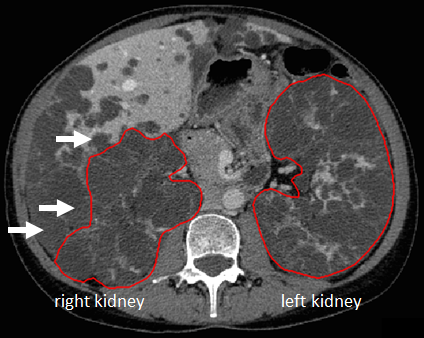}&
         \includegraphics[trim=0cm 1cm 0cm 1cm, clip=true, height=32mm]{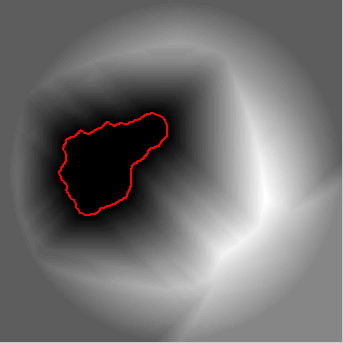}\\
          \scriptsize{(a) ADPKD Kidneys}&
          \scriptsize{(b) Geodesic Distance Map}
          \end{tabular}
	 \caption{\textbf{ADPKD Kidney Segmentation:} Panel (a) shows ADPKD kidneys. Severe morphological changes lead to difficulty in segmentation as compared to healthy kidneys. White arrows show adjacent liver cysts exhibiting similar intensity. Panel (b) shows manually segmented mid-slice along with the proposed geodesic distance map, cf. Sec. \ref{sec:proposedMethodology} for further explanation.}
	\label{fig:difficulties}
\end{figure}
\section{METHODOLOGY}
\label{sec:proposedMethodology}
As depicted in Fig. 1(a), segmenting ADPKD kidneys is a very challenging task. 
In this work we propose a semi-automatic approach for segmenting ADPKD kidneys from CT data. 
Inspired by a recent method for estimating ADPKD volume \cite{bae2013novel}, the user outlines each of the kidneys in its corresponding mid-slice, i.e., the middle slice out of all the sections containing kidney. 
Then, we compute the intensity weighted geodesic distance to the respective mid-slice segmentation at all non-segmented voxels, cf. Fig. 1(b), according to the method of \emph{Soille} \cite{soille1994generalized}. 
This yields two 3D distance volumes, one for each kidney as additional modalities (information channels). 
We demonstrate that the introduction of these novel distance volumes helps to improve the segmentation results, in contrast to a random forest classifier solely trained on the CT intensities.
Our goal is to formulate the segmentation task as a voxel-wise classification problem, where we assign to each voxel $p$ a label $\ell(p)\in\{\ell_b,\ell_r,\ell_l\}$.
While $\ell_b$ models the background class, $\ell_r$ denotes the label for the right kidney and $\ell_l$ denotes the label for the left kidney.
Based on a set of labeled examples, we aim at training a decision rule by means of a random forest classifier.
\subsection{Random Forests}
A random forest consists of a collection of decorrelated binary decision trees.
A decision tree is in turn a hierarchically ordered set of nodes, where each node has exactly $0$ or $2$ children -- in the former case, the node is called a leaf. 
Using a set of labeled examples, these decision trees are trained in order to infer the relationship between visual features and labels.
Concisely put, a random forest classifier provides a piecewise approximation of each class posterior over the feature space.
\subsection{Feature Selection}
We use so-called \textit{box features} for classification. 
As depicted in Fig. 2, a box feature at a location $p$ is defined by two offset vectors $\vec{d}_a\in\mathbb{R}^3$ and $\vec{d}_b\in\mathbb{R}^3$ which specify the centers of mass of two boxes $a$ and $b$.
These boxes are of size $x_a\times y_a \times z_a$ and $x_b\times y_b \times z_b$, respectively.
In each of these boxes we calculate the mean value w.r.t. one intensity (information) channel.
\begin{figure}[H]
	\includegraphics[width=0.5\textwidth]{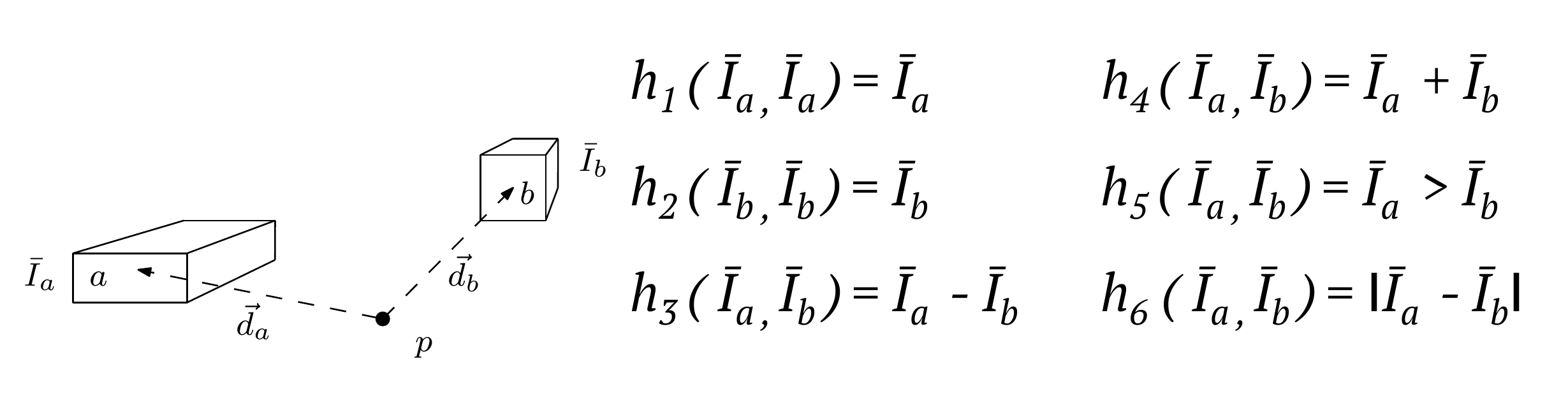}
	\caption{\textbf{Box Feature:} A box feature is defined by the two offset vectors $\vec{d}_a,\vec{d}_b\in\mathbb{R}^3$, the box sizes, and the choice of the function $h_i$ used for computing a single scalar value out of the mean values    $\bar{I}_a$ and $\bar{I}_b$. Note that these mean values could be computed from different information channels, i.e., the CT volume and the two geodesic distance volumes.}
	\label{fig:boxFeature}
\end{figure}
In particular, this means that it is possible that the mean values $\bar{I}_a$ and $\bar{I}_b$ can be computed from different channels to be able to capture inter-modality correlations. 
Once these mean values are computed, we select one of the six functions $h_j$, $j=1,\ldots,6$, cf. Fig. \ref{fig:boxFeature}, for computing a scalar feature value.
Thus, one box feature can be parametrized by a vector
\begin{equation}
(\vec{d}_a,\vec{d}_b,x_a, y_a, z_a,x_b, y_b, z_b, k_a,k_b,j),
\end{equation}
where $k_a$ and $k_b$ specify the intensity channels used for computing the respective mean value and $j=1,\ldots,6$ specifies the function used for computing the scalar feature value.
\subsection{Forest Training}
Each decision tree is trained as follows:
We randomly select a set of training samples $S$, i.e., voxels with known labels and start building the tree at the root node. 
At each node we select a splitting function defined by the selected box feature and a threshold value according to the following process:
We randomly select 100 features and compute the minimum and maximum values of the respective feature on all samples of the node.
Then we divide the range of each feature using 10 thresholds (equally spaced between the respective maximum and minimum value) and evaluate the information gain of all considered feature-threshold combinations.
The splitting function at the current node is then chosen as the combination of feature and corresponding threshold which yields the overall highest information gain. 
Although this strategy is a greedy one, it is still one of the most popular choices due to its computational efficiency \cite{criminisi2013decision}. 
This node splitting process is repeated recursively until either the maximum depth is reached, or if the number of samples sent to child nodes is too low.
Eventually, each leaf node models the class posterior estimate using a histogram from the samples that reached this leaf node.
\subsection{Forest Testing}
At testing (prediction) time, our goal is to classify newly observed voxels $p$. 
Thus, we start at the root and drive each test sample through the tree according to the splits recorded at each node during the training. 
Finally, we obtain the class histogram stored at the leaf reached by the test sample in this particular decision tree.
The overall prediction is then computed as the average of the output posteriors of each tree, and the prediction for each voxel $p$ is then given by the class with highest average posterior.
\section{EVALUATION}
\label{sec:print}
For our experiments, we used a total of 55 CT acquisitions from 41 ADPKD subjects. 
For image acquisition, a 64-slice CT scanner (LightSpeed VCT; GE Healthcare; Milwaukee, WI) was used, with single breath-hold scans and same scanning parameters for all patients (voltage 120 kV, current 150 - 500 mAs, collimation 2.5 mm, matrix 512x512, slice pitch 0.984 and increment 2.5 mm). 
All of these data sets were manually segmented by clinical experts in order to obtain ground truth annotations.
We performed a 5-fold cross-validation, i.e., we selected 44 samples for the training and tested the trained classifiers on the remaining 11 samples. 
This process was repeated five times such that every data set has been used once for validation. 
Moreover, we performed two sets of experiments. 
In the first set we trained the random forest classifiers without the geodesic distance volumes as additional information channels. 
In the second round of experiments we trained the classifier with the same settings, but with the geodesic distance volumes as additional information channels.
\begin{figure*}[htb]
\centering
\includegraphics[width=1.0\textwidth]{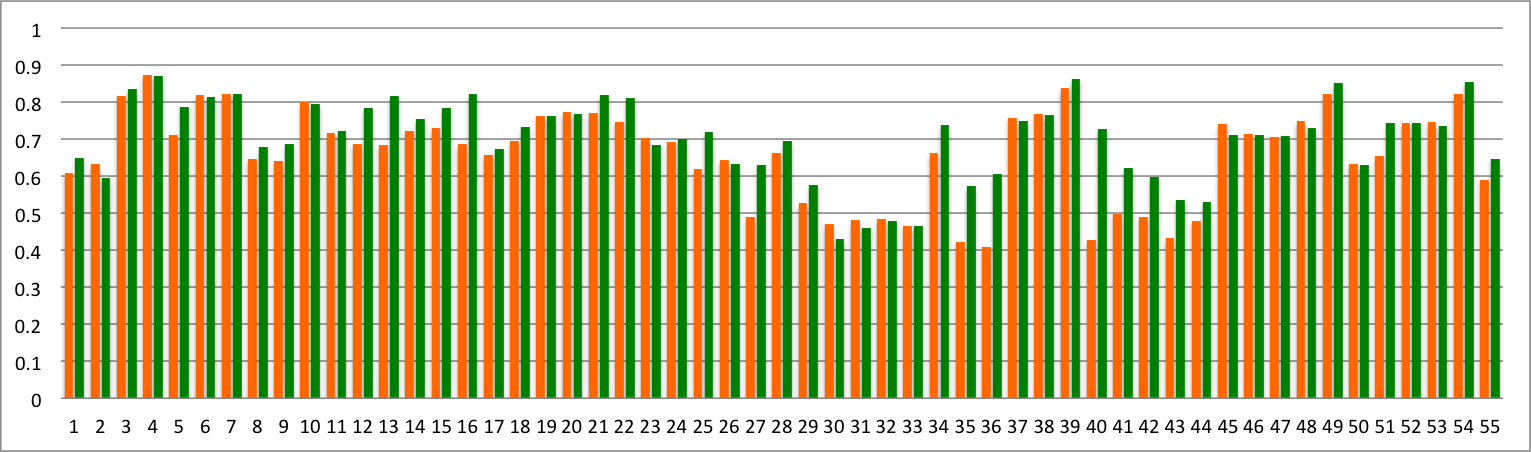}
\includegraphics[width=1.0\textwidth]{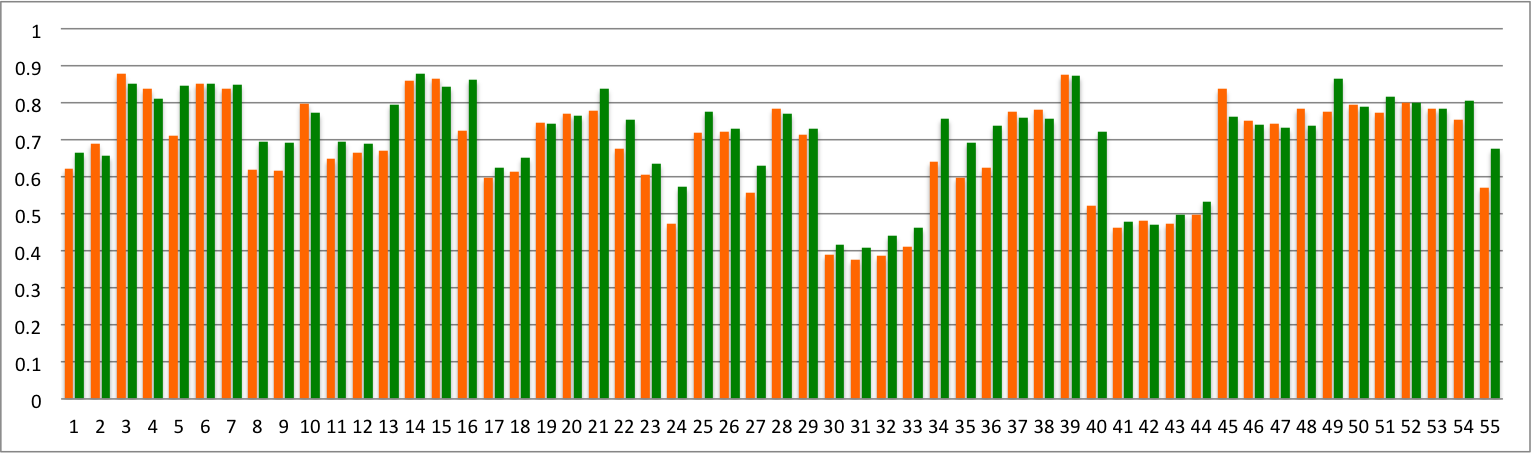}
\caption{\textbf{Dice Scores for right kidneys (upper panel) and left kidneys (lower panel)}: The results obtained for all 55 acquisitions with the baseline methods are shown in red, while the results of the proposed approach are shown in green.}
\label{fig:diceScores}
\end{figure*}
\section{RESULTS}
\label{sec:typestyle}
We computed the average dice scores for the predicted volumes for all 55 cases. 
While the green bars in Fig. 3 depict the results computed with the geodesic distance volumes, the red bars show the results for the baseline approach, i.e., the random forest classifier trained solely using the CT volumes. 
In general, we can make the following observations: Firstly, the segmentation results tend to be better for the left kidneys which is due to the fact that the boundary between the right kidney and the liver is often hard to discriminate, cf. Fig. 1. Secondly, there are a considerable number of cases where the geodesic distance volumes improve the segmentation results - especially in case of the left kidneys. 
We improved in 35 out of 55 cases by 13.25\% on average for the right kidney and in 36 out of 55 cases by 10.35\% for the left kidney, respectively, while the baseline is only better by 2\% for right kidney and by 2.42\% for the left kidney on average for the remaining 20 and 19 cases, respectively.
Therefore, the average gain of the proposed method (in the cases where it outperforms the baseline) tends to be higher than the average gain of the baseline method (in the other cases).
\section{DISCUSSION AND CONCLUSION}
\label{sec:majhead}
We presented a method for segmentation of ADPKD kidneys on contrast enhanced CT. 
Based on the results, we may draw the following two conclusions: Firstly, the proposed usage of the random forest approach helps to improve the overall segmentation process compared to baseline approach. 
Secondly, all results clearly show that segmentation of ADPKD kidneys is not at all a solved task.\\ 
The main reasons are: (i) the progressive cyst expansion in ADPKD leading to a significant and unpredictable deformation and enlargement of the kidneys, especially in patients at late stage of the disease, and (ii) the aforementioned tissue inhomogeneities of surrounding organs, for instance due to neighboring liver cysts, make fully automated kidney segmentation a challenging task.
As evaluating forests is computationally very efficient, it currently seems to be a good strategy to evaluate random forests. 
We would like to emphasize that this is to the best of our knowledge, one of the first approaches for minimally interactive segmentation of ADPKD kidneys from CT data. 
Both CT and MRI have been investigated for monitoring structural changes in ADPKD and for association between TKV and renal function or renal function decline \cite{chapman2012kidney}. 
The reason to acquire contrast enhanced CT images in the current study was to perform further renal compartment measurements.
But, these additional measurements are out of the scope of this paper.
Even at this stage, the presented method for segmentation of polycystic kidneys can potentially reduce time and effort for volumetric measurements in clinical studies. 
Moreover, we hope that this work paves way for further research regarding this challenging task. 
Future work might include the adaption of this approach to other modalities, such as MRI, or its application to other tasks, such as polycystic liver segmentation for instance and possible improvements in the current method for a more automated segmentation approach in ADPKD.

\bibliographystyle{IEEEbib}
\bibliography{refs}

\end{document}